# Adaptive Catalyst Discovery Using Multicriteria Bayesian Optimization with Representation Learning

Jie Chen[1], Pengfei Ou[2], Yuxin Chang[3], Hengrui Zhang[1], Xiao-Yan Li[2], Edward H. Sargent[2,4], & Wei Chen[1]

[1]Department of Mechanical Engineering, Northwestern University, Evanston, IL, USA
[2]Department of Chemistry, Northwestern University, Evanston, IL, USA
[3]Department of Electrical and Computer Engineering, University of Toronto, Toronto, ON, Canada
[4]Department of Electrical and Computer Engineering, Northwestern University, Evanston, IL, USA

**Abstract:** High-performance catalysts are crucial for sustainable energy conversion and human health. However, the discovery of catalysts faces challenges due to the absence of efficient approaches to navigating vast and high-dimensional structure and composition spaces. In this study, we propose a high-throughput computational catalyst screening approach integrating density functional theory (DFT) and Bayesian Optimization (BO). Within the BO framework, we propose an uncertainty-aware atomistic machine learning model, UPNet, which enables automated representation learning directly from high-dimensional catalyst structures and achieves principled uncertainty quantification. Utilizing a constrained expected improvement acquisition function, our BO framework simultaneously considers multiple evaluation criteria. Using the proposed methods, we explore catalyst discovery for the $CO_2$ reduction reaction. The results demonstrate that our approach achieves high prediction accuracy, facilitates interpretable feature extraction, and enables multicriteria design optimization, leading to significant reduction of computing power and time (10x reduction of required DFT calculations) in high-performance catalyst discovery.

**Keywords:** Catalyst screening, Bayesian Optimization, multicriteria, neural networks, uncertainty quantification.

## 1. Introduction

High-performing catalysts are crucial for sustainable energy conversion and human health. Due to huge reaction and composition spaces, the catalyst discovery cannot be achieved solely through experimental exploration. Atomistic density functional theory (DFT) simulations, aiding in understanding catalyst structures and performance, have been essential to complement experiments [1-3]. A common practice for catalyst screening is to calculate the adsorption energy using DFT and then to predict the catalyst activity and selectivity using microkinetic models. The promising candidate catalysts from simulations are then subject to experimental validation [4-8]. Nevertheless, this workflow encounters formidable challenges when employed in the high-throughput catalyst discovery, particularly in complex structure spaces, owing to the substantial computational costs of DFT [1]. Therefore, an efficient computational workflow to accelerate screening the composition−structure space is needed for high-throughput catalyst discovery.

Active learning, which adaptively queries DFT simulation, has been utilized to optimize the catalytic activity in diverse reactions such as carbon dioxide reduction [4], nitrate reduction [9], oxygen reduction [5, 10], and polymerization [11], involving a range of materials including pure metals, intermetallic compounds, binary alloys, and high-entropy alloys. A commonly used active learning method is Bayesian Optimization (BO). The fundamental concept of BO is to build a surrogate model of expensive simulations (e.g., DFT) while quantifying the model uncertainty. Next samples of simulations are determined according to an acquisition function (e.g., expected improvement [12]) that balances exploration and exploitation. Despite promising outcomes in prior attempts, current applications of BO in catalyst screening have several limitations. First, data representation is crucial to learning the structure–property relationships [13]. In existing BO frameworks, the inputs to machine learning (ML) are usually handcrafted features which are simplified summaries of chemistry and structure [11, 14-18]. However, the selection of relevant features is a difficult endeavor. Identifying and selecting features often involve a combination of trial-and-error tests and human intuition [4, 19], which are system-specific and limited for generalization [1], hindering the applications in broader materials families [20]. This calls for ML methods capable of operating directly on the high-dimensional atomic structures, thus avoiding system-specific feature engineering. Second, existing computational catalyst screening approaches are multi-step processes, i.e., narrowing down the candidates step by step considering multiple criteria such as activity, selectivity, stability, price, and toxicity, in a sequential manner. When BO approach is used for optimization, targeting solely for high activity (design objective) may result in violating other design criteria. Thus, merging the screening steps for simultaneous multicriteria evaluation is needed to enhance the effectiveness and efficiency of BO in catalyst screening.

In light of these issues, we propose a high-throughput computational catalyst screening framework with automated representation learning considering high-dimensional atomic structures and multiple evaluation criteria simultaneously. First, we develop a representation learning model, named Uncertainty-aware PointNet (UPNet), to automatically extract information from high-dimensional atomic structure data without requiring feature engineering. UPNet facilitates the creation of a ML model with principled uncertainty quantification (UQ) by taking into account the spatial correlations. Using the quantified uncertainty, our adaptive learning method effectively balances the exploration of novel catalyst structures with the exploitation of information from known optimal catalyst structures. Second, we develop a constrained BO method to enable catalyst screening considering multiple evaluation criteria simultaneously. In this case, BO searches for catalysts with high activity (i.e., desired property as an optimization objective) under the constraints of other criteria. Combining high prediction accuracy, interpretable feature extraction, and multicriteria evaluation, our method guides efficient discovery of high-performance catalysts. The success of the approach highlights the benefits of combining computational simulations and uncertainty-driven adaptive learning for high-throughput catalyst screening. Beyond catalysis, the developed method has the potential to accelerate materials discovery for various other applications [21, 22] with high-dimensional design representations and time-consuming simulations.

## 2. Results

### 2.1 Automated Feature Extraction and Bayesian Optimization for Multicriteria Catalyst Screening

The proposed Bayesian Optimization framework for catalyst screening is shown in **Fig. 1**. DFT and kinetic models [23, 24] (**Fig. 1**a and b) are used as the simulation models to calculate reactivity (e.g., activity, selectivity, and stability) for various catalyst structures. An initial set of catalyst atomic structures and calculated performance are used to build surrogate models (**Fig. 1**c). The prediction and quantified uncertainty are obtained from the surrogate models using our proposed UPNet approach. Using the constrained acquisition function (**Fig. 1**d), next samples are suggested as additional DFT simulations and added to the training set for the new BO iteration. Our work enhances the conventional BO framework to facilitate (1) automated feature extraction from the high-dimensional atomic structure by replacing the Gaussian Process model with a neural network model incorporating principled Uncertainty Quantification (**Fig. 1**c) considering spatial correlations and (2) screening with the simultaneous consideration of multiple evaluation criteria through constrained acquisition function (**Fig. 1**d).

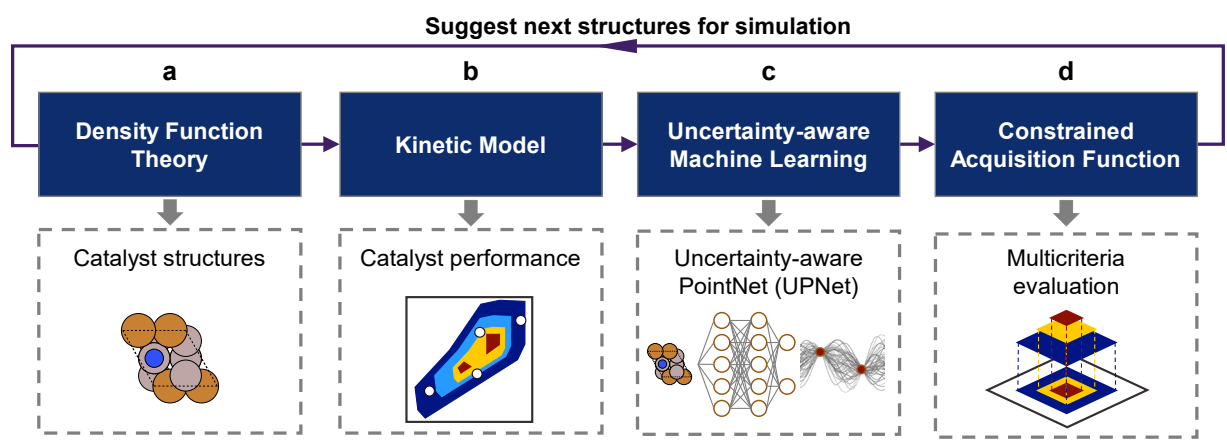

**Fig. 1 Proposed Bayesian Optimization (BO) for multicriteria catalyst screening.** The BO framework. The density functional theory (DFT) simulations and kinetic models are used for generating data for training machine learning (ML) models with principled uncertainty quantification (UQ). The constrained acquisition function suggests the next samples to query the DFT simulations considering multiple criteria.

**Fig. 2**a shows the architecture of the proposed ML surrogate model, UPNet, for prediction and principled UQ. There are two components in UPNet. First, it takes high-dimensional inputs of atomic structures used for DFT simulations, which is compiled in a tabular format shown in **Fig. 2**b. Each row of the table is one atom, represented by its element and three-dimensional locations (x, y, and z). The number of rows is the number of atoms in a molecule. In this way, the atomic structure is encoded in a point cloud format. Next, PointNet [25, 26] (**Fig. 2**c), a type of convolutional neural network model specific for point cloud data, is deployed to handle the high input dimensionality. PointNet respects the permutation invariance of points in the input. In other words, the atoms in a structure can be randomly arranged among the rows of the input table and this randomness does not lead to different predictions of the PointNet. Second, for a ML model to serve as a surrogate model in BO, it should provide principled UQ to quantify model uncertainty through distance-awareness, indicating the distance between testing data and training data (i.e., spatial correlations among the data). This is a feature of Gaussian Process [27] models but is lacking in conventional neural network models. To enable principled UQ using neural networks, Spectral-normalized Neural Gaussian Process (SNGP) [28] was developed recently with the enhanced distance-awareness capability for neural networks. In this work, we develop the UPNet approach by integrating the PointNet with the SNGP (**Fig. 2**c) through adding residual connections and spectral normalization in the hidden layers in PointNet and using the approximated Gaussian Process as the output layer. There are several other existing UQ methods in deep learning such as Bayesian neural networks and ensemble methods, but they require multiple training/inference processes and often result in high computational cost. This creates a barrier for BO since each iteration involves retraining the model and many model evaluations. SNGP can obtain prediction and UQ through a single training and inference process and is chosen for our framework due to its low memory requirement and computation cost.

To consider multiple evaluation criteria simultaneously, we formulate the screening as a constrained BO problem,

$$\min_{c(\mathbf{x}) \geq \lambda} l(\mathbf{x}). \tag{1}$$

In this case, one evaluation criterion is selected as the objective $l(\mathbf{x})$ (e.g., activity), and the remaining criteria are treated as constraints $c(\mathbf{x})$ (e.g., selectivity and stability), where $\mathbf{x}$ is the input atomic structure. $l(\mathbf{x})$ and $c(\mathbf{x})$ are both expensive to evaluate and are surrogated by regression and classification neural networks, respectively (the first and second architecture in **Fig. 2**a). By this way, the goal of the screening is to optimize the objective (finding a material targeting high activity) while not violating the constraints (e.g., selectivity or stability should exceed a threshold $\lambda$). For each iteration of the constrained BO, we select the new sample with the highest constrained expected improvement [29] for DFT simulations and then add it to the training dataset. The screening stops when a predefined total amount of iterations (i.e., computational cost) is reached or there is no further improvement of performance.

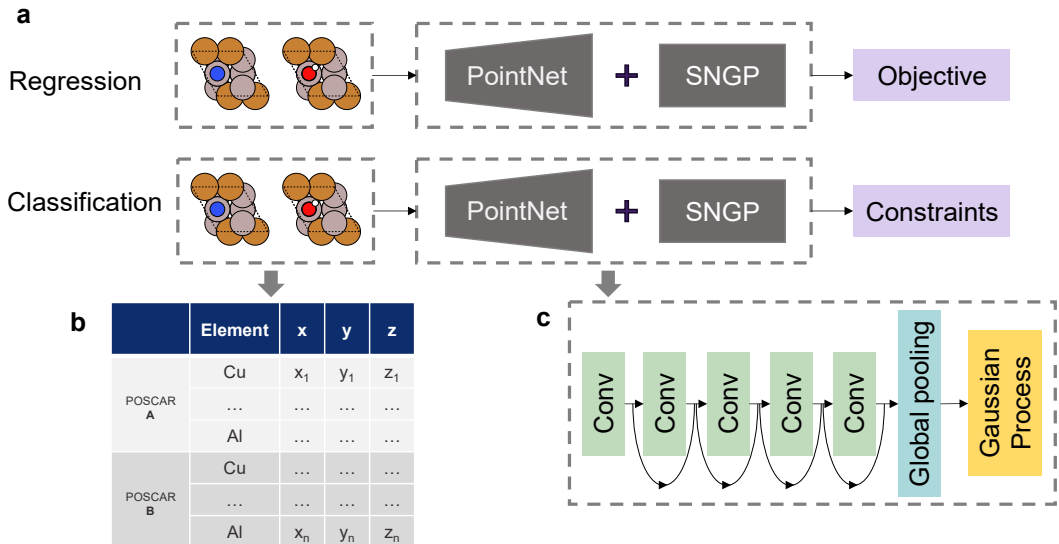

**Fig. 2 Machine learning models with principled uncertainty quantification for regression and classification. a**, The ML model as surrogate models (UPNet) in BO. Separate ML models are built for the objective (regression) and constraints

(classification). **b**, The inputs are the atomic structures formulated as point cloud data. **c**, The architecture of hidden layers and output layer of UPNet.

## 2.2 Regression and Classification with Uncertainty Quantification

For demonstrating our approach, the dataset for carbon dioxide ($CO_2$) reduction reaction is extracted from Ref. [4] where the unrelaxed structure data and adsorption energies of CO and H are provided. The unrelaxed structure data contains the atomic structures used as inputs of ML models. The adsorption energies together with the volcano scaling relationships (**Fig. 3**a and b, digitized from Ref. [30] and range normalized to [0, 1]) are used to calculate the catalyst activity and selectivity used as outputs. The distributions of the activity and selectivity are shown in **Fig. 3**c. It is noted that while the activity values are distributed relatively uniformly, the majority of selectivity is concentrated in the low-value region. Consequently, there is a high possibility of discovering a catalyst that violates selectivity requirements, which highlights the need for multicriteria screening. In Ref. [4], for the same dataset, a different active learning approach was employed for screening, utilizing only exploitation, manually engineered input features and a single evaluation criterion. In this paper, we demonstrate the benefits of using the BO approach that leverages both exploitation and exploration, by integrating automated representation learning and multicriteria evaluation to further enhance screening efficiency.

We first test the predictive performance of the UPNet model using the full dataset. Two models, one regression and one classification, are built for activity (objective) and selectivity (constraint), respectively. 0.9 is chosen as the threshold to distinguish high and low selectivity. **Fig. 3**d shows the results for the activity prediction with predicted means (points) and predictive uncertainty (±1 standard deviation, vertical bars). The mean absolute error is 11%, and the coefficient of determination ($R^2$) value is 0.69. The true values are located within one (1) predictive standard deviation. The classification results for selectivity are shown in **Fig. 3**e. The accuracy is 0.89 and the F1 score is 0.61. **Fig. 3**f shows the predictive uncertainty for selectivity classification. It can be seen that misclassifications are associated with higher uncertainty than correct classifications. We then randomly selected 10,000 samples and added random prediction errors ranging from [0, 0.2] eV to [0, 0.3] eV (with means of 0.1 and 0.15 eV, respectively) for adsorption energy, which resulted in a mean absolute error from 9% to 14% for activity and accuracy from 0.92 to 0.86 for selectivity. Thus, we conclude that the prediction errors of the developed model fall within that range based on the current dataset.

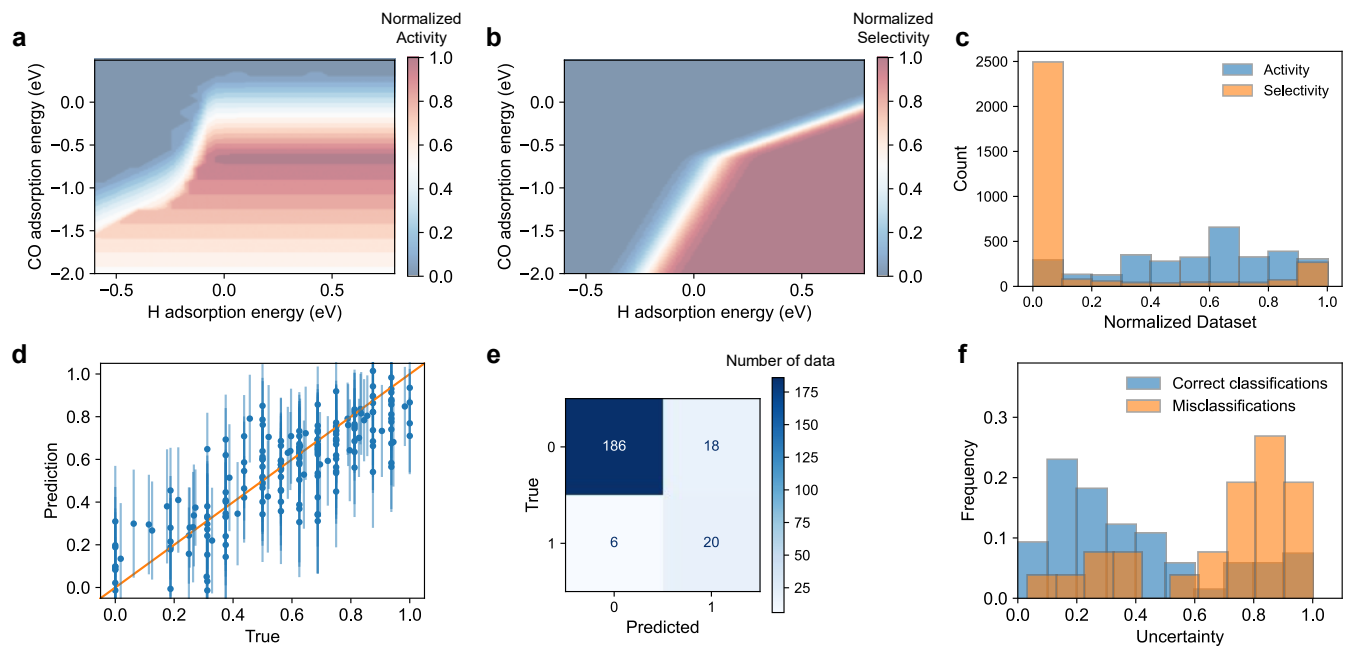

**Fig. 3 Dataset and results for regression and classification with uncertainty quantification. a**, The volcano scaling relationship for activity. **b**, The volcano scaling relationship for selectivity. Both **a** and **b** are digitized from Ref. [30] and are normalized to range [0, 1]. **a**, The distributions of activity and selectivity. **c**, The testing performance of predicting activity with uncertainty quantification (1 standard deviation). **d**, The testing performance of classifying selectivity. The uncertainty quantification for **d** is shown in **f**. Misclassifications are associated with higher uncertainty.

## 2.3 Catalyst Screening Results

After evaluating the predictive performance of the UPNet model, we use it in the proposed BO approach for catalyst screening. **Fig. 4** shows the results of BO sampling. To construct the initial data set, for each chemical composition with the number of structures greater than 10, 1 sample was selected, which resulted in an initial training dataset of 51 samples. At each iteration of BO, the sample with the highest value of constrained expected improvement was selected and added to the training dataset. In total, there are 80 iterations. The initial samples and the sequential samples from BO with and without considering constraints and random search are shown in **Fig. 4**a. The solid points indicate high selectivity (selectivity = 1) and the hollow circles indicate catalysts with selectivity smaller than 1. Overall, BO more effectively samples areas with high activity compared to random search. With the constrained BO, more solutions satisfying both high activity and high selectivity can be found compared to BO without considering constraints (**Fig. 4**a).

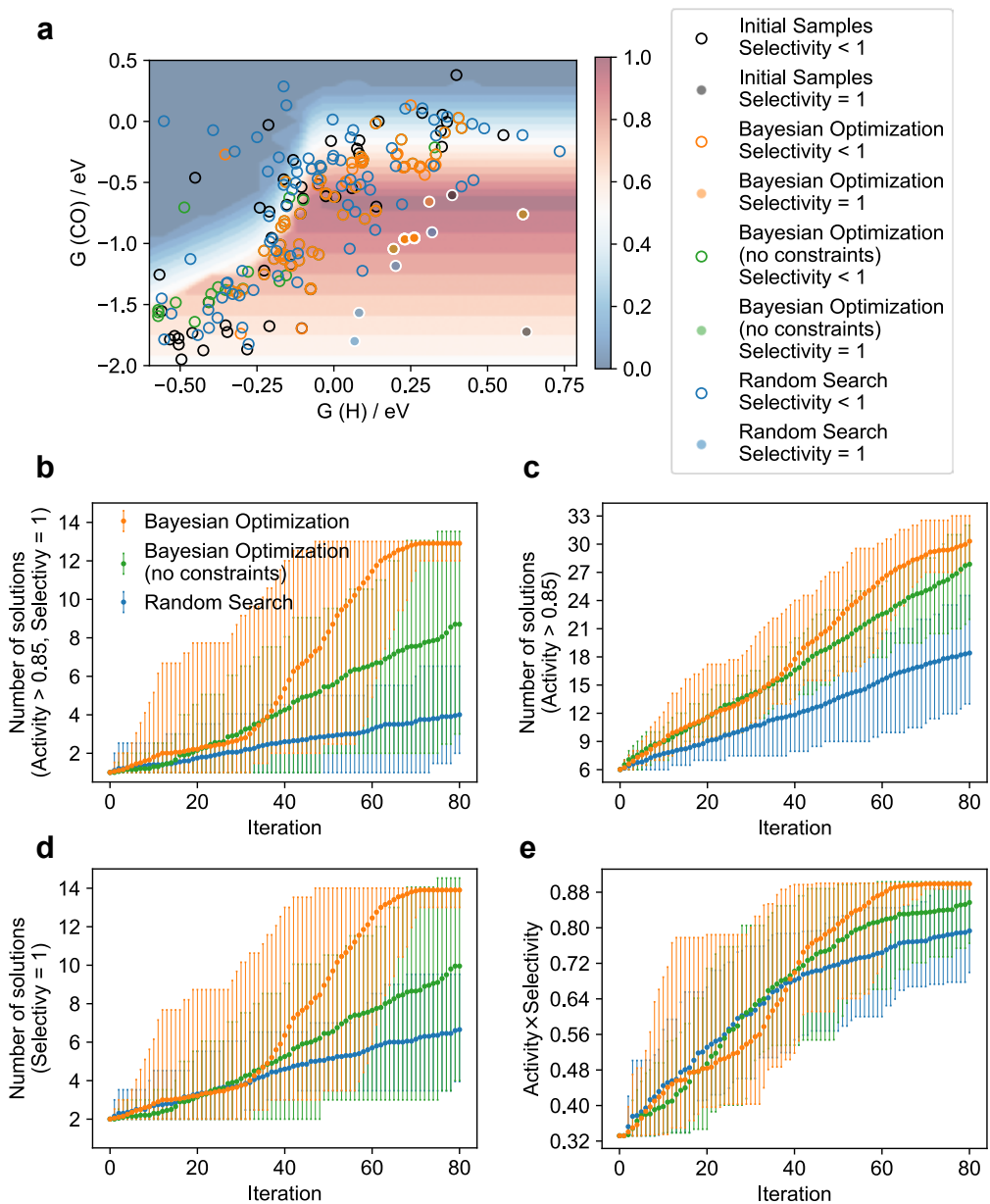

**Fig. 4 Catalyst discovery results by Bayesian Optimization with and without considering constraints and random search**. **a**, Initial samples, and samples selected by different methods. The solid points and hollow circles indicate satisfying or violating selectivity constraints. **b**, The number of solutions (high activity and high selectivity) vs. interactions. **c**, The number of solutions (high selectivity) vs. interactions. **d**, The number of solutions (high activity) vs. interactions. **e**, The average of top 10 solutions (high activity and high selectivity) vs. interactions.

To illustrate that the above results are robust against stochastic model training, the search process is repeated 20 times for each sampling method using the same initial samples. The results of means and 95% uncertainty bounds are presented in **Fig. 4**b-e. **Fig. 4**b shows the number of solutions with high activity (greater than 0.85) and high selectivity (equal to 1) for each iteration. Our proposed constrained BO method identifies more solutions than regular BO and random search. This can be further illustrated in **Fig. 4**c and d which show the solutions with higher activity and higher selectivity, respectively. The constrained BO performs comparably to unconstrained BO in optimizing the objective, characterized by high activity (**Fig. 4**c), while outperforming unconstrained BO in satisfying the constraint, namely achieving high selectivity (**Fig. 4**d). It has been observed that regular BO has a high possibility of discovering an optimal solution (activity = 1) with low selectivity when constraints are not considered. Considering the purpose of screening in identifying numerous top candidates, rather than just the single top one, for further experimental validation, we assess the screening performance of the top 10 candidates based on their average product of activity and selectivity (**Fig. 4**e)). The ideal product is equal to 1. It is noted from **Fig. 4**e that the constrained BO performs better than unconstrained BO and random search.

We also varied the number of samples per iteration (1, 2, 5, 10 and 20) when using the constrained BO. The results (SI Fig. S1) show that the convergence becomes slower with the increased number of samples per iteration. Given the same total number of samples (i.e., 80), the same performance can be achieved with less than 5 samples per iteration. Thus, with less statistical computation (i.e., more samples per iteration), more DFT calculations (i.e., more total number of samples) are needed to reach convergence of BO.

## 2.4 Latent Space Interpretability

A desired feature of a surrogate model used in BO is proper spatial correlation modeling, i.e., the capability to quantify the distance among data. To visualize the spatial distance, the t-distributed stochastic neighbor embedding (t-SNE) method is used to map the high-dimensional data to low-dimensional space (two-dimensional in this work) while preserving significant data structure (i.e., pairwise similarities) [31]. The key idea is that similar data in the high-dimensional space are closer to each other in the low-dimensional space. The data of chemical compositions each with more than 25 structures are extracted and used as training data for SNGP. **Fig. 5**c shows the embeddings of latent space (the features from the last hidden layer before the output layer) from SNGP (also in SI Fig. S2). The black points in **Fig. 5**c represent out-of-distribution data which have only one structure per chemical composition. It can be observed that the same chemical compositions are clustered and separated among the clusters and most of the out-of-distribution data is separated from the training data. This shows the capability of UPNet to preserve the similarities of the catalyst structures from the input space to the latent space, which is necessary for differentiating seen and unseen data. Nevertheless, it is noted that there exists another cluster of out-of-distribution data that is close to the training data (a mixture of different chemical compositions). One plausible explanation is that the SNGP model discerns challenging-to-predict data, mapping them closer in the latent space. Next, the distributions of the standard deviations which are the outputs of SNGP are plotted in **Fig. 5**d. One data point per chemical composition is partitioned from the training dataset to serve as in-distribution testing data. It can be seen that the uncertainty for in-distribution testing data is within the range of training data and the uncertainty for out-of-distribution data is distinctively higher than training data and in-distribution testing data, which is a desired property of the surrogate model in BO to predict high/low uncertainty for dissimilar/similar unseen data to the seen data (or far away from/close to the seen data in terms of the spatial distribution). For comparison with the latent space obtained from the proposed UPNet (**Fig. 5**c), the latent space from mere PointNet (CNN model) without the capability of uncertainty quantification is visualized in **Fig. 5**b. In this case, catalyst structures with different chemical compositions can be mapped close together in the latent space (i.e., feature collapses) for training data, and the out-of-distribution data are distributed among and close to the training data, which show the spatial distance from the input space is not well preserved without integrating PointNet with SNGP (UPNet in **Fig. 5**c). **Fig. 5**a shows the t-SNE of the input space (i.e., the catalyst structures). Due to the high dimensionality of inputs, it is difficult for t-SNE to separate out-of-distribution data from the training data, which pinpoints the effectiveness of the extracted low dimensional features [32-34] in the latent space of UPNet.

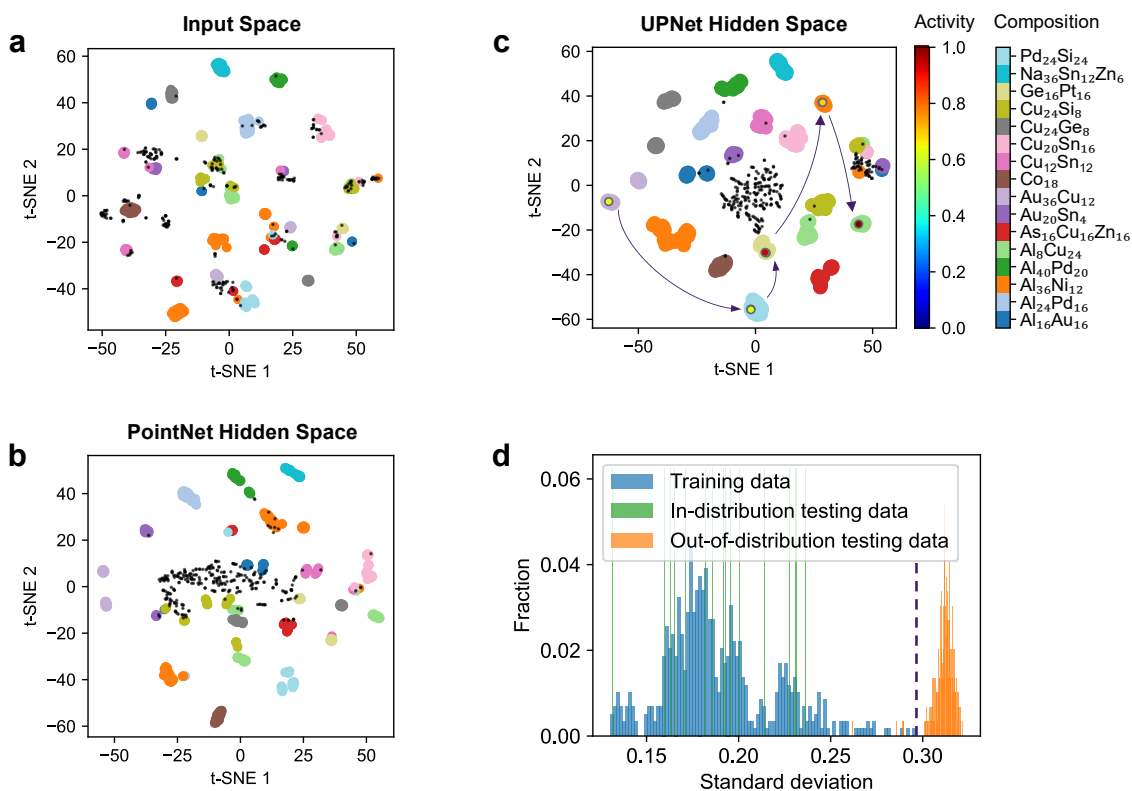

**Fig. 5 t-Distributed stochastic neighbor embedding (t-SNE) and uncertainty quantification. a**, t-SNE of latent space from Spectral-normalized Neural Gaussian Process (SNGP) model. **b**, t-SNE of latent space from CNN model. **c**, t-SNE of the input space. **d**, Uncertainty (standard deviation) distribution of training data, in-distribution testing data, and out-of-distribution testing data from the SNGP model. **e**, t-SNE of the search path from constrained Bayesian Optimization.

Furthermore, we tracked the search path of BO, and the first 15 sequential samples are illustrated in the low dimensional latent space (**Fig. 5**c). It is noted that the search traverses among 5 clusters of chemical compositions. Before jumping to another cluster (i.e., exploration), multiple samples were selected within one cluster (i.e., exploitation). This shows that BO balances exploration and exploitation to find the global optimum. The activity (design objective) increases throughout the search process. $Al_8Cu_{24}$ was selected by the proposed constrained BO approach, which is consistent with the conclusion [30] that Cu-Al alloys exhibiting a higher proportion of Cu than Al are of great potential for $CO_2$ reduction. Compared to the existing active learning approach [30], the total number of data is reduced from approximately 4,000 to 131 (51 initial training data and 80 during constrained BO), showing the efficiency of our proposed approach.

## 3. Conclusions

In summary, we develop an adaptive catalyst discovery framework combining DFT and active learning. Due to the high computational cost of DFT and vast varieties of catalyst structures and compositions, BO is utilized to adaptively sample the catalyst space. To accelerate screening, we seek to predict catalytic properties directly from the atomic structures, which avoids problem-specific feature engineering, and to search for the catalyst with high activity while satisfying other evaluation criteria (e.g., high selectivity) simultaneously. To achieve these two goals, we extended the BO framework by developing an uncertainty-aware physics-based ML model (UPNet) and utilizing a constrained expected improvement acquisition function, respectively. With UPNet, the high-dimensional atomic coordinates and descriptors of the catalysts can be directly used as the input, which are physically meaningful, generic, and easy to obtain, in contrast to handcrafted features. Furthermore, the UPNet can reliably quantify the uncertainties of the predictions, which enables exploration and exploitation of the vast catalyst space. Next, we formulate multicriteria screening as a constrained optimization problem with the activity as the objective and other evaluation criteria as constraints. The constrained acquisition function augments the expected improvement function with the probability of constraint satisfaction. We demonstrated the method in the discovery of $CO_2$ reduction reaction catalyst, a promising material for pursuing carbon neutrality. The results show that the UPNet model archives high prediction accuracy. The model can also extract interpretable latent features due to its ability to preserve the similarities of the catalyst structures from the input space to the latent space. The developed constrained BO method outperforms unconstrained BO and random search in terms of the quantity and quality of the discovered catalysts with top performance. With the generic representation of high dimensional catalyst structures, our proposed methods can seamlessly extend to catalyst discovery in more complex structure spaces (e.g., high-entropy alloy), and for other reactions [35]. In the broad context of materials research, this work provides a general method for automated multicriteria screening, facilitating high-throughput materials discovery.

## 4. Materials and Methods

### 4.1 Data Preparation

We sourced our datasets from the publicly available datasets https://github.com/ulissigroup/GASpy_manuscript [4], which contained two separate sets: one with 20,909 intermetallic-CO adsorption datapoints and the other with 22,675 intermetallic-H adsorption datapoints. We paired the datapoints in each set that shared the same intermetallic surface and same adsorption position. We also excluded the data outside the range of the activity/selectivity map in **Fig. 3**a and b. This resulted in a CO-H paired dataset with 3,163 datapoints. The dataset contains the unrelaxed intermetallic structures adsorbing CO and H, and adsorption energies of CO and H. The activity and selectivity were added to the dataset after calculation using adsorption energies and activity/selectivity map.

### 4.2 Bayesian Optimization

Bayesian Optimization (BO) is a ML-based optimization method [36] solving the problem

$$\min f(\mathbf{x}), \tag{2}$$

where is **x** the design variable. The objective function $f(\mathbf{x})$ is expensive to query in terms of computational, monetary, or opportunity cost. Typically, $f(\mathbf{x})$ lacks the property of convexity or linearity and can only be observed without information on derivatives, which makes it difficult to optimize using gradient-based methods. The goal of BO is to find global optimum instead of local optima.

BO involves two main components: a surrogate model of the objective function for statistical inference, and an acquisition function to decide the next samples to query $f(\mathbf{x})$. BO is an iterative optimization process: First, initial samples are obtained according to the design of experiments. A surrogate model is then built using the observed samples. The predictions and corresponding uncertainties are used to construct the acquisition function. Next, new samples are observed which maximize the acquisition function and are added to the dataset. Following that, the surrogate model is updated with the augmented dataset and the iteration continues until a predefined target is reached or budget is exhausted.

The subsequent sections discuss the above two main components of BO customized in this work for catalyst screening. First, a typical surrogate model for $f(\mathbf{x})$ is Gaussian Process which requires the dimension of **x** to be not too large. This paper overcomes the dimensionality restriction by developing a neural network-based model (UPNet). Second, BO was initially developed for unconstrained optimization. To enable multicriteria catalyst screening, a constrained acquisition function is adopted in this paper.

## 4.3 Uncertainty Quantification for Deep Learning

Denote a data generation distribution $p(y, \mathbf{x}) = p(y|\mathbf{x})p(\mathbf{x})$, where $\mathbf{x} \in \mathcal{X}$ is the input and $y$ is the output. We can express the conditional data generation distribution $p(y|\mathbf{x})$ [37, 38] as

$$p(y|\mathbf{x}) = p(y|\mathbf{x}, \mathbf{x} \in \mathcal{X}_{IND}) \times p(\mathbf{x} \in \mathcal{X}_{IND}) + p(y|\mathbf{x}, \mathbf{x} \in \mathcal{X}_{OOD}) \times p(\mathbf{x} \in \mathcal{X}_{OOD}), \quad (3)$$

where $\mathcal{X}_{IND}$ and $\mathcal{X}_{OOD}$ are in-domain and out-of-domain data, respectively. To formulate the uncertainty quantification problem as a learning problem, a proper loss function needs to be defined to evaluate the quality of the predictive uncertainty from the model [39-41]. In Ref. [28], the loss function of expected risk over the entire input domain $\mathcal{X}$ is used to construct an optimal predictive distribution. As a result, $p(y|\mathbf{x})$ is expressed as

$$p(y|\mathbf{x}) = p(y|\mathbf{x}, \mathbf{x} \in \mathcal{X}_{IND}) \times p(\mathbf{x} \in \mathcal{X}_{IND}) + p_{uniform}(y|\mathbf{x}, \mathbf{x} \in \mathcal{X}_{OOD}) \times p(\mathbf{x} \in \mathcal{X}_{OOD}). \quad (4)$$

which can be interpreted as follows: if a data point is in the training data domain $\mathcal{X}_{IND}$ (i.e., high $p(\mathbf{x} \in \mathcal{X}_{IND})$), the predictive distribution from the model is trusted, otherwise, a uniform distribution is assigned representing a lack of knowledge. Thus, it is crucial to estimate $p(\mathbf{x} \in \mathcal{X}_{IND})$ for reliable uncertainty quantification in ML, which requires the model to be distance-aware to quantify the similarity between a data point and the in-domain training data.

Spectral-normalized Neural Gaussian Process (SNGP) was developed [28] to quantify uncertainty in deep learning through distance-awareness. SNGP replaces the commonly used dense output layer with an approximated Gaussian Process (GP) layer which models the distance from the last hidden layer to the output layer. To handle the large dimensionality and large data size, Laplace-approximation inference is applied to the random Fourier features expansion of the GP layer [42]. This operation enables the model to be distance-aware from the last hidden layer to the output layer. There is a feature collapse issue in neural networks. That is, the testing data which is dissimilar to the training data can be possibly mapped close to the training data in the hidden representation space [43]. Thus, it is needed to preserve the distance from the input layer throughout the hidden layers so that the distance in the hidden layers $\|h(\mathbf{x}_1) - h(\mathbf{x}_2)\|$ to reflect the distance in the input layer $\|\mathbf{x}_1 - \mathbf{x}_2\|$. SNGP preserves the distance approximately through adopting a residual neural network architecture and up-bounding the spectral norm of the weights in the nonlinear residual blocks.

The SNGP outputs predictive logits that follow Gaussian distribution $N(\mu(\mathbf{x}), \sigma^2(\mathbf{x}))$. For regression, the predictive mean and variance are $\mu(\mathbf{x})$ and $\sigma^2(\mathbf{x})$, respectively. For classification, the predictive probability [44] is

$$p(y|\mathbf{x}) = \text{softmax}\left(\frac{\mu(\mathbf{x})}{\sqrt{1 + \pi/8 \cdot \sigma^2(\mathbf{x})}}\right). \quad (5)$$

The demonstrations of using SNGP for regression and classification are shown in SI Fig. S3.

**4.4 Machine Learning Model Architecture**

The architecture of the ML model is illustrated in **Fig. 2**a, and the detailed specifications are provided in Table 1. The output is activity or selectivity. The input is the catalyst structure. As shown in **Fig. 2**b, the structural information is encapsulated as a matrix. Each row represents an atom. The columns represent the features of the atoms which are obtained without the need to perform DFT simulations, containing one-hot encoding of the chemical element, atomic mass, electronegativity, atomic radius, and the coordinates of the three-dimensional location. The input catalyst structure is a format of point cloud data since each atom corresponds to a point. We used the PointNet model to handle the point cloud data. PointNet is a type of convolutional neural network (CNN). Different from the regular CNN, the convolutional layers are connected sequentially without pooling layers in between. At the end of all convolutional layers, there is a global pooling layer. The convolutional layers have 64 filters and a kernel size of 1 by 45 (the number of atom features). This architecture provides permutation invariance, that is, the random arrangement of the rows in the input matrix does not influence the model training and outputs. This work integrates PointNet with SNGP (Table 1). We added residual connections between adjacent convolutional layers, i.e.,

$$x_{l+1} = F(x_l) + x_l, \quad (6)$$

where $x_l$ and $F(x_l)$ are the input and the output of the $l$-th layer, respectively. Strides of size 1 and "same" padding are added to keep the dimension of the filters unchanged after each convolutional operation to enable the summation in Eq. (6). The spectral normalization with spectral norm bound equal to 0.95 is applied in each convolutional layer. The output layer is a Random Feature Gaussian Process [28]. The model is trained using Adam optimizer [45] with the mean squared error (for regression) or categorical cross entropy (for classification) loss function, learning rate of 1e-5, batch size of 32, and epochs of 500 for regression or 300 for classification.

Table 1 UPNet layers and specifications.

| Layers | Specifications |
|---|---|
| Input Layer | Shape: (147, 45) |
| Convolutional layer 1 | Kernel size: 1 × 45, Filter size: 64, Activation: ReLU, Strides = 1, Padding = same, Spectral normalization, Spectral norm bound = 0.95 |
| Residual layer 1 | Input Layer + Convolutional layer 1 |
| Convolutional layer 2 | Same specifications as Convolutional layer 1 |
| Residual layer 2 | Residual layer 1 + Convolutional layer 2 |
| Convolutional layer 3 | Same specifications as Convolutional layer 1 |
| Residual layer 3 | Residual layer 2 + Convolutional layer 3 |
| Convolutional layer 4 | Same specifications as Convolutional layer 1 |
| Residual layer 4 | Residual layer 3 + Convolutional layer 4 |
| Convolutional layer 5 | Same specifications as Convolutional layer 1 |
| Residual layer 5 | Residual layer 4 + Convolutional layer 5 |
| Pooling layer | Global max pooling |
| Output layer | Random Feature Gaussian Process, Size: 1 for regression and 2 for classification, Activation: ReLU |

### 4.5 Constrained Acquisition Function

In this work, the constraint $c(\mathbf{x}) \geq \lambda$ in Eq. (1) is treated as a classification problem. That is, the label is 1 if $c(\mathbf{x}) - \lambda \geq 0$ and 0 otherwise. The constrained BO is fulfilled through a constrained acquisition function, i.e., constrained expected improvement [29],

$$EI_C(\hat{\mathbf{x}}) = Pr[c(\hat{\mathbf{x}}) - \lambda \geq 0] \, EI(\hat{\mathbf{x}}). \tag{7}$$

$EI(\hat{\mathbf{x}})$ is the expected improvement for a candidate sample $\hat{\mathbf{x}}$,

$$EI(\hat{\mathbf{x}}) = (\mu(\hat{\mathbf{x}}) - f_t^{max})\Phi\left(\frac{\mu(\hat{\mathbf{x}}) - f_t^{max}}{\sigma(\hat{\mathbf{x}})}\right) + \sigma(\hat{\mathbf{x}})\phi\left(\frac{\mu(\hat{\mathbf{x}}) - f_t^{max}}{\sigma(\hat{\mathbf{x}})}\right), \tag{8}$$

where $f_t^{max}$ is the maximum value observed till iteration $t$, $\Phi(\cdot)$ is the standard Gaussian cumulative distribution function, and $\phi(\cdot)$ is the standard Gaussian probability density function. $Pr[c(\hat{\mathbf{x}}) - \lambda \geq 0]$ is the probability of the constraint being satisfied, i.e., $p(y = 1|\hat{\mathbf{x}})$ according to Eq. (5).

## Acknowledgements

The project described was supported by National Science Foundation (NSF) DMR-2219489, the Center for Hierarchical Materials Design under Award No. CHiMaD NIST 70NANB19H005, and the International Institute for Nanotechnology (IIN). H.Z. was supported by the Ryan Graduate Fellowship.

Research was sponsored by the Army Research Laboratory and was accomplished under Cooperative Agreement Number W911NF-22-0121. The views and conclusions contained in this document are those of the authors and should not be interpreted as representing the official policies, either expressed or implied, of the Army Research Laboratory of the U.S. Government. The U.S. Government is authorized to reproduce and distribute reprints for Government purposes notwithstanding any copyright notation herein

The authors would like to express their appreciation to Dr. Daniel W. Apley, Dr. Vinayak P. Dravid, Dr. Roberto dos Reis, Wei Liu, Carolin B. Wahl, Jiezhong Wu, and Alfred Yan for valuable discussions and comments.

**Author Contributions.** J.C., P.O., and H.Z. conceived the idea. Y.C., J.C., P.O., and X.L. prepared the data. J.C. developed and coded the UPNet model and constrained Bayesian Optimization algorithm. J.C. and Y.C. wrote the manuscript. E.H.S., and W.C. supervised the project. All authors discussed the results and commented on the manuscript.

**Competing Interest Statement.** The authors declare no competing interests.

**Data and code availability.** The dataset (adsorption energy, activity and selectivity maps, etc.) and the code for constrained Bayesian Optimization using UPNet are available at https://github.com/jcj7292/Multicriteria-Catalyst-Discovery-Using-Automated-Feature-Extraction-Bayesian-Optimization.

**Supporting Information for**
Expedited Catalyst Discovery Using Multicriteria Bayesian Optimization with Automated Representation Learning

Jie Chen[1], Pengfei Ou[2], Yuxin Chang[3], Hengrui Zhang[1], Xiao-Yan Li[2], Edward H. Sargent[2,4], & Wei Chen[1]

[1]Department of Mechanical Engineering, Northwestern University, Evanston, IL, USA
[2]Department of Chemistry, Northwestern University, Evanston, IL, USA
[3]Department of Electrical and Computer Engineering, University of Toronto, Toronto, Ontario, Canada
[4]Department of Electrical and Computer Engineering, Northwestern University, Evanston, IL, USA

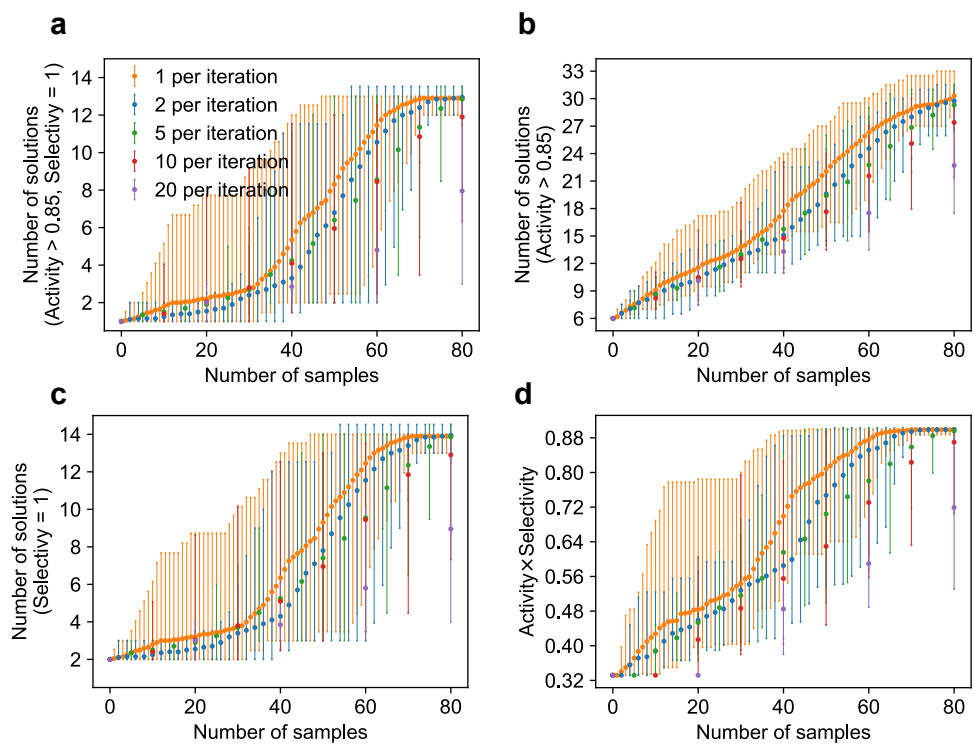

**Fig. S1 Impact of number of samples for iteration for constrained Bayesian Optimization**. **a**, The number of solutions (high activity and high selectivity) vs. the number of samples. **b**, The number of solutions (high selectivity) vs. the number of samples. **c**, The number of solutions (high activity) vs. the number of samples. **d**, The average of top 10 solutions (high activity and high selectivity) vs. the number of samples.

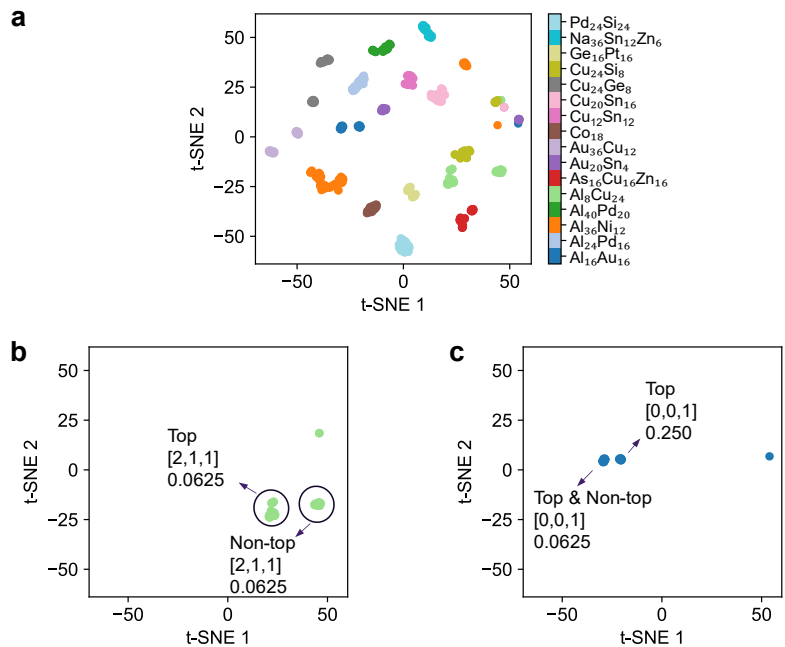

**Fig. S2 Examples of latent space clusters**. **a**, t-SNE of latent space from Spectral-normalized Neural Gaussian Process (SNGP) model. **b** and **c**, The demonstration examples for $Al_8Cu_{24}$ and $Al_{16}Au_{16}$, respectively. The three rows of notation near the clusters are top, miller index, shift, respectively, where top (and non-top) indicates the chosen surface was at the top (or bottom) of the originally enumerated surface; miller index is a 3-tuple of integers indicating the Miller indices of the surface; shift represents for c-direction shift used to determine cutoff for the surface [1]. This suggests that these three types of information influence the similarities among subclusters.

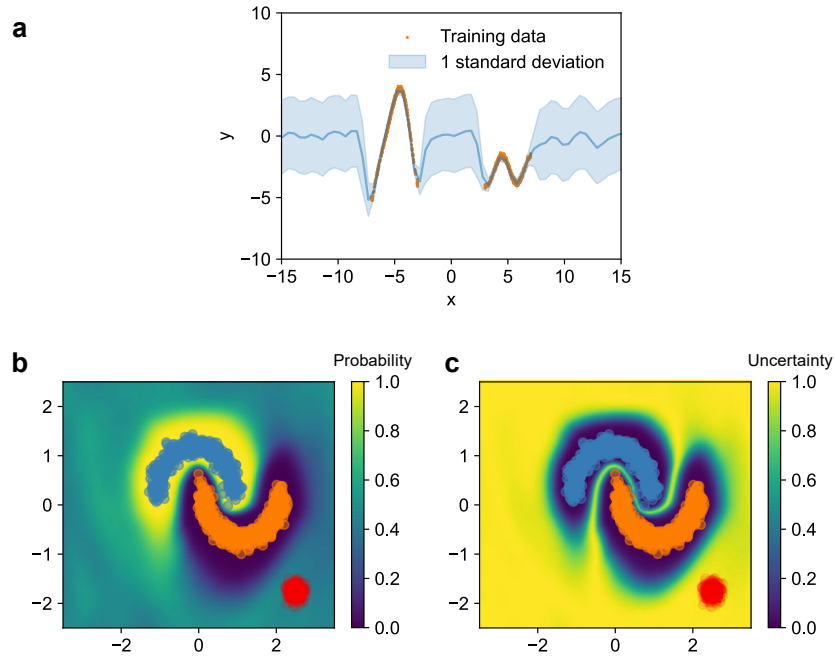

**Fig. S3 Demonstration of Spectral-normalized Neural Gaussian Process (SNGP)**. In order to apply SNGP with PointNet, this demonstration is to show the effectiveness of SNGP with convolution layers. **a**, A regression demonstration. The input of the regression problem is a 10 by 2 matrix with repeated values. The mean predictions match the training data. The predictive uncertainty (i.e., the standard deviation) is small (or large) near (or far away from) the training data. **b** and **c**, A classification demonstration. The input size is 10 by 2 with each row representing the location of a point and repeated rows. The blue and orange points are the binary training data. The red points are the out-of-distribution data. Results in **b** show that the probability near the training data is close to 0 or 1 and is 0.5 far away from the training data. Correspondingly, in **c**, the uncertainty is low (close to 0) near the training data and is high for the out-of-distribution data.